\def\year{2022}\relax
\documentclass[letterpaper]{article} %
\usepackage{aaai22}  %
\usepackage{times}  %
\usepackage{helvet}  %
\usepackage{courier}  %
\usepackage[hyphens]{url}  %
\usepackage{graphicx} %
\usepackage{natbib}  %
\usepackage{caption} %
\DeclareCaptionStyle{ruled}{labelfont=normalfont,labelsep=colon,strut=off} %
\usepackage{booktabs}
\usepackage[table,xcdraw]{xcolor}
\usepackage{multirow}
\usepackage{subcaption}
\usepackage{algorithm}
\usepackage{algorithmic}
\usepackage{amsmath}

\usepackage{newfloat}
\usepackage{listings}
\floatstyle{ruled}
\newfloat{listing}{tb}{lst}{}
\floatname{listing}{Listing}
\usepackage{color}%

\title{Explain, Edit, and Understand: Rethinking User Study Design \\ for Evaluating Model Explanations}
\author {
    Siddhant Arora\thanks{denotes equal contribution.}\textsuperscript{\rm 1} \quad 
    Danish Pruthi\footnotemark[1]\textsuperscript{\rm 1} \quad 
    Norman Sadeh\textsuperscript{\rm 1}  \\ 
    William W. Cohen\textsuperscript{\rm 2} \quad 
    Zachary C. Lipton\textsuperscript{\rm 1} \quad 
    Graham Neubig\textsuperscript{\rm 1}
}
\affiliations {
    \textsuperscript{\rm 1} Carnegie Mellon University \\
    \textsuperscript{\rm 2} Google AI \\
    \{siddhana, ddanish, sadeh, zlipton, gneubig\}@cs.cmu.edu, wcohen@google.com
}

\usepackage{bibentry}

\begin{document}

\maketitle

\begin{abstract}
    In attempts to ``explain''  predictions
    of machine learning models,
    researchers have proposed 
    hundreds of techniques
    for attributing predictions 
    to features that are 
    deemed
    important.
    While these attributions are often claimed
    to hold the potential to improve 
    human ``understanding'' of the models,
    surprisingly little work explicitly 
    evaluates progress towards this aspiration. 
    In this paper, we conduct a crowdsourcing study, 
    where participants interact with deception detection models
    that have been trained to distinguish 
    between genuine and fake hotel reviews. 
    They are challenged both 
    to simulate the model on fresh reviews,
    and to edit reviews with the goal 
    of lowering the probability 
    of the originally predicted class.
    Successful manipulations would lead to an adversarial example.
    During the training (but not the test) phase, %
    input spans are highlighted to communicate salience.
    Through our evaluation,
    we observe that for a linear bag-of-words model,
    participants with access to the feature coefficients 
    during training are able to 
    cause a larger reduction in model confidence in the testing phase when 
    compared to the no-explanation control. 
    For the BERT-based classifier, 
    popular \emph{local explanations} do not improve 
    their ability to reduce the model confidence over the no-explanation case. 
    Remarkably, when the explanation for the BERT model
    is given by the (global) attributions 
    of a linear model trained to imitate the BERT model,
    people can effectively manipulate the model.\footnote{The code used for our study is available at: \url{https://github.com/siddhu001/Evaluating-Explanations}.}
\end{abstract}

\section{Introduction}
\begin{figure*}[ht]
    \centering
    \includegraphics[width=0.92\linewidth]{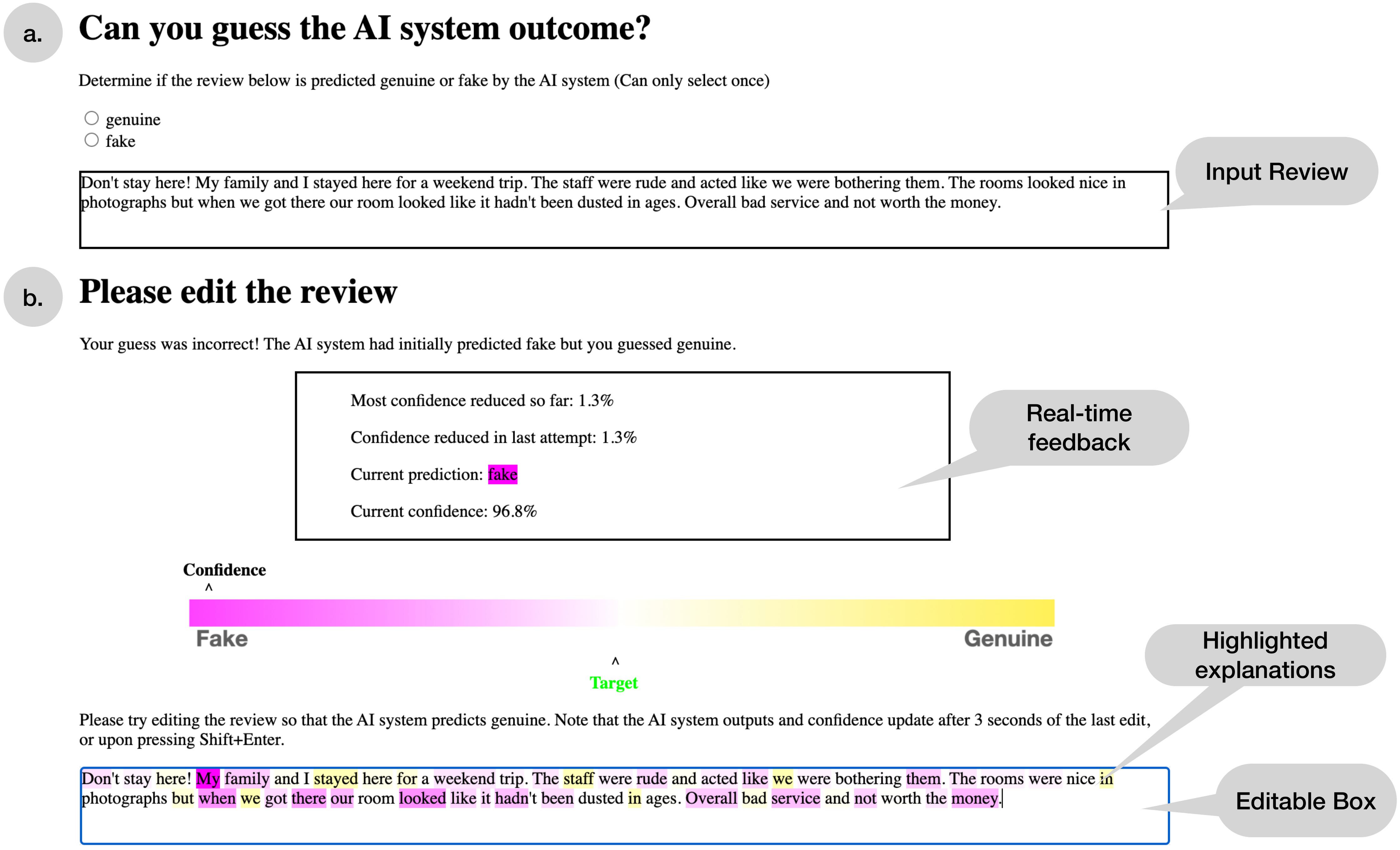}
  \caption{Our user study, 
  as shown to participants during 
  the training phase: 
  a) first, participants guess the model prediction; 
  (b) next, they edit the review 
  to reduce the model confidence
  towards the predicted class.
  Through highlights, 
  we indicate the attribution scores 
  produced by different techniques. 
  Participants receive feedback on their edits,
  observing updated predictions, 
  confidence and attributions, all in real time. 
  The test phase does not include attributions 
  but is otherwise similar to the training.}
  \label{fig:overall_flow}
\end{figure*}

Owing to their remarkable predictive accuracy
on supervised learning problems,
deep learning models are increasingly  
deployed in consequential domains,
such as medicine~\cite{medimg1, medimg2}, 
and criminal justice~\cite{criminal1}.
Frustrated by the difficulty of communicating
what precisely these models have learned,
a large body of research has sprung up
proposing methods that are purported 
to \emph{explain} their predictions~\cite{doshi2017towards, lipton2018mythos, guidotti2018survey}.
Typically, these so-called explanations
take the form of saliency maps,
attributing the prediction %
to a subset of the input features,
or assigning weights to the features
according to their salience.
To date, while hundreds of such attribution techniques
have been proposed~\cite{ribeiro2016should,visualexp, sundararajan2017axiomatic, smilkov2017smoothgrad}, 
what precisely it means for a feature to be salient
remains a point of conceptual ambiguity.
Thus, many proposed techniques 
are evaluated only via 
visual inspection of a few examples 
where the highlighted features agree
with the author's (and reader's) intuitions.
Across  papers, one common motivation 
for such attributions is to improve 
human ``understanding'' of the models~\cite{ribeiro2016should, doshi2017towards, sundararajan2017axiomatic}. 
However, whether these attributions 
confer understanding 
is seldom evaluated explicitly 
and there is relatively little work that 
characterizes what explanations enable people to do. 

One suggestion to evaluate model understanding
is to use \emph{simulatability} 
as a proxy for understanding---i.e.,
if a participant can accurately predict
the output of the model on unseen examples
~\cite{doshi2017towards}. 
Following this idea,
a few prior studies 
examine if explanations
help humans predict model output~\citep{chandrasekaran2018explanations, hase-bansal-2020-evaluating}. 
Such studies are typically divided 
into a training and a test phase. 
In the training phase, participants see
a few input, output, explanations triples, 
and in the test phase, they are asked 
to guess the model output 
for unseen examples.\footnote{
\citet{chandrasekaran2018explanations}
present model explanations during the testing phase,
whereas \citet{hase-bansal-2020-evaluating}
do not include explanations at test time, 
as explanations could ``leak'' model output
(see~\citet{pruthi2020evaluating,jacovi2020aligning}.).} 
Many prior studies on evaluating model explanations 
have reached negative results,
noting that they do not definitively 
aid humans 
in predicting model behavior on 
visual question answering~\cite{chandrasekaran2018explanations}  
and text classification tasks~\cite{hase-bansal-2020-evaluating}.

In this paper, we rethink the user design 
for evaluating model explanations
for text classification tasks, 
and propose two key changes.    
First, we provide participants with 
\textbf{query access to the model}: 
they can alter input documents 
to observe how model predictions
and explanations change in real time.
Second, we extend the simulation task 
by prompting participants \textbf{to edit examples 
to reduce the model confidence} 
towards the predicted class.
While prior work \cite{kaushik2019learning} prompts
humans to edit examples for data augmentation, editing exercises haven’t been explored for evaluating explanations. 
This editing exercise allows 
us to capture detailed metrics, 
e.g., average confidence reduced, 
which, as we shall later see,
can be used to compare 
relative utility of 
different explanation techniques. 

We perform a crowdsourcing study 
using the proposed paradigm
on a deception detection task, 
with machine learning models that are trained
to detect whether hotel reviews 
are genuine or fake~\cite{ott2011finding}.
In this task, the human performance  
is only slightly better 
than that of random guessing,
while machine learning models 
are significantly more accurate, 
making it an interesting testbed 
for studying whether attributions
help people to understand the 
associations employed by the models.
In our study, the participant first guesses 
whether the given hotel review
is classified as fake or genuine
(see Figure~\ref{fig:overall_flow}).
We then prompt the participant
to edit the review 
such that the model confidence 
towards the predicted class is reduced.
During the training phase, 
we present attributions 
by highlighting input spans.
For instance, the attribution 
in Figure~\ref{fig:overall_flow} 
suggests that the model associates 
tokens ``My'' and ``family'' 
with the fake class, 
perhaps indicating that 
fake reviews tend to mention external factors
instead of details about the hotel.
In our setup, a participant could 
test any such hypotheses, 
by editing the example 
and observing the updated predictions,
outputs, and attributions immediately.

Through our study, we seek to answer the question: 
Which (if any) attribution techniques
improve humans' ability to guess the model output,
or edit the input examples 
to lower the model confidence?
From the evaluation methodology standpoint, 
we assess if the interactive environment
with query access to the models 
makes it possible to distinguish 
the relative value of different attributions.
For these research questions, we compare 
popular attribution techniques---LIME \cite{ribeiro2016should} 
and integrated gradients \cite{sundararajan2017axiomatic},
against a no-explanation control.

Our evaluation reveals that (i)~for both a 
linear bag-of-words model 
and a BERT-based classifier, 
none of the explanation methods
definitively help participants to simulate 
the model's output more accurately 
at test time (when explanations are unavailable); 
(ii)~however, access to feature coefficients 
from a linear model during training 
enables participants to cause a larger reduction
in the model confidence at test time; 
and (iii)~most remarkably, %
feature coefficients and global cue words\footnote{
  Words that correspond 
  to the largest feature coefficients.
}
 from a 
linear (student) model trained 
to mimic a (teacher) BERT model
significantly help participants 
to manipulate BERT.
Additionally, we notice that participants respond 
to the highlighted spans, as over $40\%$ of 
all the edits  %
are performed on these spans. 
Our comparisons
lead to 
quantitative
differences 
among evaluated attributions, 
underscoring the efficacy  
of our paradigm.

\section{Related Work}
We briefly discuss past attempts to 
evaluate explanation methods, 
both via user studies and automated metrics.
\paragraph{Simulatability-based Evaluation.} 

Model simulatability measures   
human ability to 
predict the model output on fresh examples.  
It is a prominent %
metric to evaluate explanation methods, 
and 
is treated as a proxy for model understanding~\cite{doshi2017towards}.
Using simulatability, a recent study evaluates 
five different explanation generation schemes 
for text and tabular classification tasks~\cite{hase-bansal-2020-evaluating}.
Their study runs two different types of tests: 
(i) \emph{forward simulation} which measures
simulatability on unseen examples without explanations, 
after presenting participants $16$ 
training examples with explanations; and (ii) \emph{counterfactual simulation}
which captures participants' ability to guess the model output of perturbed 
input examples while 
observing the true labels, predictions, and explanation 
for the original examples.
The study concludes that for the text classification task, 
none of the five evaluated explanations 
definitively help 
participants better simulate the model in the forward simulation task 
(when explanations were provided only at training time). 
The participants in their study report 
found it difficult to retain the insights learned 
from the training phase during the testing phase.
Another 
study 
examines 
the extent to which explanations from a VQA model
help humans predict its responses and failures~\cite{chandrasekaran2018explanations}.
In their setup, visual saliency maps were provided 
both during the training and the testing phase. 
This study too leads to a similar conclusion: visual attributions 
do not help in simulating the VQA model. 
Another recent study 
measures simulatability of 
several regression models 
that estimate the value of real-estate listings~\cite{poursabzi2021manipulating}. 
They observe that participants could simulate a 
linear model with $2$ features 
but fail to simulate one with $8$ features. 
They also note that participants 
could not correct model mistakes for any of the given models. 
Another paper 
investigates 
if humans could predict model output using explanations alone,
and found erasure and attention-based explanations to be useful~\cite{treviso2020towards}.  

Our work differs 
with the above studies in a number of ways: 
none of the prior studies allow participants 
to test out models 
for inputs of their choice 
(query access). 
Additionally, we ask participants to edit examples 
with a goal to reduce the model confidence, 
in an attempt to identify adversarial examples. 
This exercise allows us to capture 
detailed metrics, including the average 
amount of confidence reduced and number of examples 
successfully flipped. Furthermore, 
we interleave the training and test phase thereby 
mitigating retention issues reported in ~\cite{hase-bansal-2020-evaluating}.

\paragraph{Other User Studies.}
There have been 
several crowdsourcing studies 
that evaluate different  
aspects of explanations~\cite{userstudyRel1,userstudyRel2,userstudyRel3,userstudyRel5,
userstudyRel4,userstudyRel6}.
~\citet{mohseni2021multidisciplinary} categorize
these efforts based
on the goals they aim to achieve, the intended audience,  
 and the evaluation metrics. 
Few studies measure 
if explanations enable 
participants to better predict the task output (i.e., the ground truth)
instead of the model output---specifically, 
if explanations 
help participants 
gain sufficient insights to 
distinguish genuine reviews from fake reviews~\citep{lai2019human,lai2020deceptive}. 
~\citet{lertvittayakumjorn2019human}
evaluate
if explanations 
help in identifying 
the better performing model.
Lastly, a recent study 
examines saliency maps for an age-prediction model, 
and concludes that none of the
 explanations impact human's trust in the model~\cite{andreas2020visual}.

\paragraph{Automated Metrics.} A variety of  
automated metrics %
to measure explanation quality have been proposed in the past. 
However, many of them can be easily
gamed~\cite{hookergame1,trevisogame2,hasegame3} (see~\cite{pruthi2020evaluating} for a detailed discussion on this point). 
A popular way to evaluate explanations is to 
compare the produced explanations with expert-collected rationales~\cite{rationale1,rationale2}. 
Such metrics only capture 
whether the produced explanations are plausible, 
but do not comment upon the \emph{faithfulness} of explanations 
to the process through which predictions are obtained. A 
recently proposed approach 
quantifies the value of explanations %
via the accuracy gains that they confer on a student model trained
to simulate a teacher model~\cite{pruthi2020evaluating}. 
Designing automated evaluation metrics %
is an ongoing area of research, 
and to the best of our knowledge,
none of the automated metrics have been demonstrated 
to correlate with 
any human measure of explanation quality. %

\section{Evaluation through Iterative Editing}
\label{Experiment}

This section first describes our evaluation paradigm 
and discusses how it is different from prior efforts. 
We then introduce several metrics 
for evaluating model explanations. 

\subsection{Experimental Procedure}
We divide our evaluation into two alternating phases: 
a training phase and a test phase. 
During the training phase, participants 
first read the input example, 
and are challenged to guess the model prediction.
Once they submit their guess, 
they see the model output, 
model confidence and an explanation 
(explanation type varies across treatment groups).
As noted earlier, 
several prior studies
solely evaluate model simulatability~\cite{hase-bansal-2020-evaluating, chandrasekaran2018explanations, treviso2020towards, poursabzi2021manipulating}. 
We extend the past protocols and 
further prompt participants to edit the 
input text with a goal to lower
the confidence of the model prediction.
As participants edit the input, they see 
the updated predictions, 
confidence and explanations 
in real time (see Figure~\ref{fig:overall_flow}).
Therefore, they can validate
any hypothesis about the %
input-output associations (captured by the model), 
by simply editing the text 
based on their hypothesis and observing 
if the model prediction changes accordingly.
The editing task concludes 
if the participants are able to flip the model prediction, 
or run out of the allocated time 
(of three minutes).
The instructions for the study prohibit participants 
to edit examples in a manner that changes the
meaning of the 
text (more details in the next section).

The test phase is similar to the training phase 
except for an important distinction: 
explanations are not available during testing %
so that we can evaluate the insights participants 
have acquired 
without the support of explanations. 
Holding back the explanations at test time 
eliminates concerns that
the explanations might trivially 
leak the output 
(see ~\citet{pruthi2020evaluating} for a detailed discussions).
In our study, after every two examples in the training stage, 
participants complete one test example. 
In contrast to past studies,
where participants first review all the 
training examples before attempting the test examples,
we show participants one test example 
after every two training examples.
In the~\citet{hase-bansal-2020-evaluating} study,
participants report that it was difficult 
to retain insights from the training phase 
during the later test round. 
Our interleaving procedure alleviates such concerns.

\subsection{Metrics}
While simulatability has been used
as a proxy measure for model
understanding~\cite{hase-bansal-2020-evaluating, chandrasekaran2018explanations}, 
we argue that simulating the model 
is a difficult task for people, 
especially after viewing only a few examples.
Hence, we propose to compute detailed metrics 
that are based on participants' ability to edit the example 
to lower the model confidence towards predicted class 
and to possibly flip model predictions.
We believe that such metrics 
are finer-grained indicators 
for participant's understanding of the model, 
since participants might not comprehend 
how different factors 
combine to produce the output, 
but they may identify 
a few input-output associations,
which they can effectively apply 
in the manipulation exercise.

Based on this motivation,
we measure three metrics
(a)~the simulation accuracy,
(b)~average reduction in model confidence,
and (c)~percentage of examples flipped.
Following prior work~\cite{andreas2020visual}, 
we use mixed effects regression models  
to estimate these three quantities.
For each experiment, 
a participant is randomly placed 
in one of the $5$ cohorts.
All participants in the same cohort 
see the same training and test examples, 
irrespective of the experiment.
Further, across different cohorts, 
test examples differ (but we use a fixed set of examples for training). 
We use multiple cohorts so as to not  
rely on 
a few test examples for our conclusions.
The mixed effects models include fixed effect term
$\beta_{\text{treatment}}$ for each treatment
and a random effect intercept $\alpha_{\text{cohort}}$ to
determine the impact of the cohort
to which a participant is assigned. %
Since mixture effect models 
can effectively handle random variability 
introduced due to different data samples  
and different participant cohorts, 
it is an appropriate choice 
to isolate the impact 
of each explanation type.
The three mixed effects models 
can be described as
\begin{align*} 
\mathbf{y}_{\text{target}} = \beta_{0} + \beta_{\text{treatment}} \times x_{\text{treatment}}+ \alpha_{\text{cohort}} \times x_{\text{cohort}},
\end{align*}
where the target corresponds
to three evaluation metrics discussed above 
and $\beta_{0}$ is the intercept.

\section{A Case Study of Deception Detection}
\label{Task}

We choose a deception detection task---distinguishing between fake and real hotel reviews \cite{ott2011finding}---as the backdrop for our 
crowdsourcing study.
This is because prior studies have noted 
that humans struggle  
with this task while 
machine learning models %
are significantly more accurate.  
Our motivation for using this setup is 
that models 
exploit  
subtle, unknown and possibly counter-intuitive 
associations to drive prediction, 
providing an interesting testbed to evaluate 
whether attributions communicate such associations.
Further, since human accuracy is low for this task,
the participants do not have preconceived notions that could
potentially conflate with the
 simulation task. 
Therefore, 
this task
makes an interesting testbed 
to characterize how much explanations 
help humans in understanding 
the input-output associations that deception detection models exploit. 
The study comprises $20$ training, 
and $10$  testing examples in total, 
and lasts for $90$ minutes per participant.
\begin{table}[h]
    \centering
  {\small
  \begin{tabular}{lcc}
  \toprule
  Model & Accuracy \\ \midrule
  Human Accuracy ~\cite{ott2011finding}  & $\approx$ 60\% \\ %
  Logistic Regression & 87.8\% \\ 
  BERT  & 89.8\% \\
  \bottomrule
  \end{tabular}
  }
  \caption{Accuracy on the deception detection task.}
  \label{tab:accuracy-ML-models}
  \end{table}

\begin{table*}[t]
    \centering
  \begin{tabular}{ll|l|lll}
  \toprule
  \rowcolor[HTML]{FFFFFF} 
  Model                                                         & Treatments                                                                                               & \begin{tabular}[c]{@{}c@{}} Simulation \\Accuracy \end{tabular} & Phase  & \begin{tabular}[c]{@{}c@{}} Examples flipped \\ (Percentage) \end{tabular}                                                            & \begin{tabular}[c]{@{}c@{}} Avg. Confidence \\ Reduced \end{tabular}                                                  \\ 
  \midrule
  \rowcolor[HTML]{FFFFFF} 
  \cellcolor[HTML]{FFFFFF}                                      & \cellcolor[HTML]{FFFFFF}       &                                                                               & Train                                              & {\color[HTML]{1D1C1D} \hphantom{0}8.2 \small{{[}\hphantom{0}5.4, 11.6{]}}}                               & \hphantom{0}8.0 \small{{[}\hphantom{0}7.0, \hphantom{0}9.0{]}}                                                \\  
  \cellcolor[HTML]{FFFFFF}                                      & \multirow{-2}{*}{\cellcolor[HTML]{FFFFFF}Control }   & \multirow{-2}{*}{\cellcolor[HTML]{FFFFFF}53.1 \small{[50.0, 57.0]} }     &  \cellcolor[HTML]{e0e0e0}Test                      & \cellcolor[HTML]{e0e0e0}{\color[HTML]{1D1C1D} 15.0 \small{{[}10.8, 19.4{]}}}    & \cellcolor[HTML]{e0e0e0}{\color[HTML]{1D1C1D} \hphantom{0}5.9 \small{{[}\hphantom{0}4.3, \hphantom{0}7.8{]}}} \\
  \rowcolor[HTML]{FFFFFF}
  \cellcolor[HTML]{FFFFFF}                                      & \cellcolor[HTML]{FFFFFF}                                                             &                         & Train                                & {\color[HTML]{1D1C1D} \bf{36.7 \small{{[}24.8, 49.3{]}}}}                         & {\color[HTML]{1D1C1D} \bf{21.3 \small{{[}19.5, 23.1{]}}} }                  \\  
  \multirow{-4}{*}{\begin{tabular}[l]{@{}l@{}} Logistic \\ Regression \end{tabular}} & \multirow{-2}{*}{\cellcolor[HTML]{FFFFFF}Feature coefficients} & \multirow{-2}{*}{\cellcolor[HTML]{FFFFFF}54.5 \small{[51.0, 58.0]}  }  & \cellcolor[HTML]{e0e0e0}Test &  \cellcolor[HTML]{e0e0e0}{\color[HTML]{1D1C1D} 16.0 \small{{[}10.8, 21.6{]}} } & \cellcolor[HTML]{e0e0e0}\bf{\hphantom{0}8.9 \small{{[}\hphantom{0}7.2, 10.6{]}}}                 \\ 
  \midrule
  \rowcolor[HTML]{FFFFFF} 
  \cellcolor[HTML]{FFFFFF}                                      & \cellcolor[HTML]{FFFFFF}                                                                        &               & Train                                             & 15.0 \small{{[}11.6, 18.8{]}}                                                    & {\color[HTML]{1D1C1D} 10.7 \small{{[}\hphantom{0}8.6, 12.8{]}}}                        \\  
  \rowcolor[HTML]{e0e0e0} 
  \cellcolor[HTML]{FFFFFF}                                      & \multirow{-2}{*}{\cellcolor[HTML]{FFFFFF}Control }          & \multirow{-2}{*}{\cellcolor[HTML]{FFFFFF}57.1 \small{[54.0, 61.0]} }                              & Test                                                & {\color[HTML]{1D1C1D} 12.4 \small{{[}\hphantom{0}7.6, 18.1{]}}}                              & {\color[HTML]{1D1C1D} \hphantom{0}9.2 \small{{[}\hphantom{0}6.6, 11.9{]}}}                          \\
  \rowcolor[HTML]{FFFFFF} 
  \cellcolor[HTML]{FFFFFF}                                      & \cellcolor[HTML]{FFFFFF}                                                                               &              & Train                  & {\color[HTML]{1D1C1D} 14.4 \small{{[}10.5, 19.5{]}}}                             & {\color[HTML]{1D1C1D} 10.2 \small{{[}\hphantom{0}8.2, 12.3{]}}}                    \\
  \rowcolor[HTML]{e0e0e0} 
  \cellcolor[HTML]{FFFFFF}                                      & \multirow{-2}{*}{\cellcolor[HTML]{FFFFFF}LIME }           & \multirow{-2}{*}{\cellcolor[HTML]{FFFFFF}56.4 \small{[53.0, 60.0]} }                                   & Test                      & {\color[HTML]{1D1C1D} \hphantom{0}7.7 \small{{[}\hphantom{0}4.4, 11.3{]}}}                               & {\color[HTML]{1D1C1D} \hphantom{0}6.1 \small{{[}\hphantom{0}4.1, \hphantom{0}8.2{]}}}                      \\ 
  \rowcolor[HTML]{FFFFFF} 
  \cellcolor[HTML]{FFFFFF}                                      & \cellcolor[HTML]{FFFFFF}                                                                               &    & Train                                            & {\color[HTML]{1D1C1D} \bf{23.6 \small{{[}19.4, 28.0{]}}}}                       & {\color[HTML]{1D1C1D} \bf{16.5 \small{{[}14.0, 19.2{]}}}}                \\  
  \rowcolor[HTML]{e0e0e0} 
  \cellcolor[HTML]{FFFFFF}                                      & \multirow{-2}{*}{\cellcolor[HTML]{FFFFFF}Integrated gradients }                 & \multirow{-2}{*}{\cellcolor[HTML]{FFFFFF}56.6 \small{[54.0, 60.0]} }                      & Test                                              & {\color[HTML]{1D1C1D} 13.6 \small{{[}\hphantom{0}8.2, 19.3{]}}}                            & {\color[HTML]{1D1C1D} 10.4 \small{{[}\hphantom{0}7.7, 13.3{]}}}                    \\  [-1.75ex]
   & \cline{1-5} 
  \rowcolor[HTML]{FFFFFF} 
  \cellcolor[HTML]{FFFFFF}                                      & \cellcolor[HTML]{FFFFFF}                                                                           &        & Train                                                & {\color[HTML]{1D1C1D} \bf{32.2 \small{{[}27.1, 37.3{]}}}}                         & {\color[HTML]{1D1C1D} \bf{22.6 \small{{[}19.7, 25.6{]}}}}               \\ 
  \rowcolor[HTML]{e0e0e0} 
   \cellcolor[HTML]{FFFFFF}                                       & \multirow{-2}{*}{\cellcolor[HTML]{FFFFFF}{\begin{tabular}[l]{@{}l@{}} \cellcolor[HTML]{FFFFFF}{Feature coefficients} \\ \cellcolor[HTML]{FFFFFF}{\small{(from a linear student)}} \end{tabular}}}             & \multirow{-2}{*}{\cellcolor[HTML]{FFFFFF}60.5 \small{[57.0, 64.0]} }               & Test                                           & {\color[HTML]{1D1C1D} \bf{21.3 \small{{[}15.7, 27.4{]}}}}                         & {\color[HTML]{1D1C1D} \bf{14.9 \small{{[}11.6, 18.4{]}}} }               \\
  \rowcolor[HTML]{FFFFFF} 
  \cellcolor[HTML]{FFFFFF}                                      & \cellcolor[HTML]{FFFFFF}                                                                       &          & Train                                              & {\color[HTML]{1D1C1D} \bf{40.6 \small{{[}32.0, 49.6{]}}}}                         & {\color[HTML]{1D1C1D} \bf{29.9 \small{{[}26.8, 33.0{]}}}}               \\ 
  \rowcolor[HTML]{e0e0e0} 
  \multirow{-10}{*}{\cellcolor[HTML]{FFFFFF}BERT}                & \multirow{-2}{*}{\cellcolor[HTML]{FFFFFF}{\begin{tabular}[l]{@{}l@{}} \cellcolor[HTML]{FFFFFF}{\quad $+$ global cues } \\ \cellcolor[HTML]{FFFFFF}{\small{(from a linear student)}} \end{tabular}}}            & \multirow{-2}{*}{\cellcolor[HTML]{FFFFFF}55.7 \small{[51.0, 60.0]} }                  & Test                                          & {\color[HTML]{1D1C1D} \bf{31.6 \small{{[}23.2, 40.8{]}}}}                         & {\color[HTML]{1D1C1D} \bf{23.6 \small{{[}19.7, 27.6{]}}} }               \\ 
  
  \bottomrule
  \end{tabular}
  \caption{We report human performance across
    different explanations in our study. 
  None of the explanations help participants to
  simulate the models, %
  whereas global explanations for the BERT model and 
  feature coefficients for the logistic regression model help to  
  reduce model confidence.
  Bold values indicate statistically significant differences 
  as compared to the no-explanation control (p-value $<0.05$). %
  Square brackets indicate bootstrapped 95\% confidence intervals. %
The simulation accuracy is computed together 
as participants see the explanations 
only after guessing the model predictions 
in both the train and test phase. 
  }
  \label{tbl:overall_results}
  \end{table*}

\paragraph{What are permissible edits?} 
We ask participants not to alter the staying experience 
conveyed through the hotel review. 
If the review is positive, negative or mixed,
then the edited version should 
maintain that stance. 
They are allowed to paraphrase
and can remove or change information not relevant 
to the experience about the hotel. 
For instance, changing ``My husband and I'' to ``We''
is valid edit.
However, inventing details that 
influence the experience about the hotel are not permitted (e.g.,~adding ``The staff was unfriendly'' is not allowed).
To enforce these guidelines, 
we (1) discard submissions 
where the edit distances
between the original and edited version is large%
\footnote{We remove submissions where the word edit distance $>0.9$ of the length of input review, or if half of original words are deleted.}
and then (2) manually inspect the edits to reject submissions 
that violate our instructions.

\subsection{Machine Learning Models}

We consider two
machine learning models for our experiments. 
The first is a linear logistic regression model with unigram TF-IDF features. 
The second model is a BERT-based classification model~\cite{devlin-etal-2019-bert}.
We train, or finetune,
these models using the deception review dataset~\cite{ott2011finding}.
We use the original 
train/validation/test splits,
which are class balanced (i.e., exactly half of the reviews are genuine). 
For the logistic regression model,
we select hyperparameters, i.e.~regularization strength and regularization penalty,
via a $10$-fold cross-validation,
whereas we use the default parameters of the BERT model.
The accuracy of the two models is significantly 
higher than the estimated human performance on this task, which is 
around $60\%$ (Table~\ref{tab:accuracy-ML-models}). 
We refer readers to ~\cite{ott2011finding} 
for details on the dataset and estimating human performance for this task.

\subsection{Controls \& Treatments}
Participants are randomly placed 
into different 
control and treatment groups which vary based on
the type of explanations offered and the choice of  
the machine learning model.
For both the linear logistic regression model and the BERT model,
we run a control study without explanations. 
For the linear model, we use 
feature coefficients of unigram features 
as explanations in the treatment group.
For the BERT model,
we use the following explanation-based treatments.%
\paragraph{Local Explanations.} 
Local explanation refer to techniques that
produce explanations by 
observing how the model's predictions change
upon perturbing the input slightly 
For the BERT classifier, 
we experiment with two widely-used 
local explanations: 
LIME~\cite{ribeiro2016should} and integrated gradients~\cite{sundararajan2017axiomatic}.
LIME produces an explanation using the feature coefficients of 
a linear interpretable 
model that is trained
to approximate the original model
in the local neighborhood 
of the input example. 
Integrated gradients are computed by integrating gradients 
of the log-likelihood of the predicted label 
along the line joining a 
starting reference point and the given input example.
These explanations are 
presented to participants through highlights~(see Figure~\ref{fig:overall_flow}). 
 \begin{table*}[]
    \centering
  \begin{tabular}{ll|l|lll}
  \toprule
  \cellcolor[HTML]{FFFFFF}Model                                 & \cellcolor[HTML]{FFFFFF}Treatments                                                         & \begin{tabular}[c]{@{}c@{}} Simulation \\Accuracy \end{tabular} & Phase  & \begin{tabular}[c]{@{}c@{}} Examples flipped \\ (Percentage) \end{tabular}                                                            & \begin{tabular}[c]{@{}c@{}} Avg. Confidence \\ Reduced \end{tabular}                                                  \\
  \midrule
  \rowcolor[HTML]{FFFFFF} 
  \cellcolor[HTML]{FFFFFF}                                      & \cellcolor[HTML]{FFFFFF}                                                              & \cellcolor[HTML]{FFFFFF}                                                              &Train                         & \cellcolor[HTML]{F8F8F8}{\color[HTML]{1D1C1D} \bf{\hphantom{-}29.3 \small{{[}\hphantom{-}16.5, 42.1{]}}}} & \cellcolor[HTML]{F8F8F8}{\color[HTML]{1D1C1D} \bf{\hphantom{-}13.8 \small{{[}\hphantom{-0}7.6,  19.9{]}}}}  \\  
  \rowcolor[HTML]{e0e0e0} 
  \multirow{-2}{*}{\cellcolor[HTML]{FFFFFF}{\begin{tabular}[l]{@{}l@{}} \cellcolor[HTML]{FFFFFF}{Logistic} \\ \cellcolor[HTML]{FFFFFF}Regression \end{tabular}}} & \multirow{-2}{*}{\cellcolor[HTML]{FFFFFF}$\beta_{\text{~Feature coefficients}}$} & \multirow{-2}{*}{\cellcolor[HTML]{FFFFFF}\hphantom{-}2.3 {[}-2.1, 6.7{]}} & Test                          & {\color[HTML]{1D1C1D} \hphantom{-0}2.6 \small{{[}\hphantom{0}-3.2, \hphantom{0}8.4{]}}}                           & {\color[HTML]{1D1C1D} \hphantom{-0}2.9 \small{{[}\hphantom{0}-0.2, \hphantom{0}6.0{]}} *}                        \\
  \midrule
  \rowcolor[HTML]{FFFFFF} 
  \cellcolor[HTML]{FFFFFF}                                      & \cellcolor[HTML]{FFFFFF} & \cellcolor[HTML]{FFFFFF}                                                              & Train                         & \cellcolor[HTML]{F8F8F8}{\color[HTML]{1D1C1D} \hphantom{0}-0.5 \small{{[}\hphantom{0}-8.6, \hphantom{0}7.6{]}}}  & \cellcolor[HTML]{F8F8F8}{\color[HTML]{1D1C1D} \hphantom{0}-0.3 \small{{[}\hphantom{0}-6.4,   \hphantom{0}5.7{]}}} \\  
  \rowcolor[HTML]{e0e0e0} 
  \cellcolor[HTML]{FFFFFF}                                      & \multirow{-2}{*}{\cellcolor[HTML]{FFFFFF}$\beta_{~\text{LIME}}$}   & \multirow{-2}{*}{\cellcolor[HTML]{FFFFFF}-0.0 {[}-5.5, 5.4{]}}             & Test                          & {\color[HTML]{1D1C1D} \hphantom{0}-4.4 \small{{[}-12.5, \hphantom{0}3.7{]}}}                         & {\color[HTML]{1D1C1D} \hphantom{0}-2.9 \small{{[}\hphantom{0}-8.6, \hphantom{0}2.8{]}}}                         \\  
  \cellcolor[HTML]{FFFFFF}                                      & \cellcolor[HTML]{FFFFFF}   & \cellcolor[HTML]{FFFFFF}                                                           & \cellcolor[HTML]{FFFFFF}Train  & \cellcolor[HTML]{FFFFFF}{\color[HTML]{1D1C1D} \bf{\hphantom{-0}8.8 \small{{[}\hphantom{-0}0.6, 16.9{]}}}}  & \bf{\hphantom{-0}6.7 \small{{[}\hphantom{-0}0.7, 12.6{]}}}                                                   \\  
  \rowcolor[HTML]{e0e0e0} 
  \cellcolor[HTML]{FFFFFF}                                      & \multirow{-2}{*}{\cellcolor[HTML]{FFFFFF}$\beta_{~\text{Integrated gradients}}$} & \multirow{-2}{*}{\cellcolor[HTML]{FFFFFF}-1.2 {[}-9.9, 3.1{]}}             & Test                          & {\color[HTML]{1D1C1D} \hphantom{-0}1.2 \small{{[}\hphantom{0}-6.7, \hphantom{0}9.1{]}}}                           & \hphantom{-0}1.0 \small{{[}\hphantom{0}-4.6, \hphantom{0}6.6{]}}                                                   \\  
[-1.75ex]
  & \cline{1-5}
  \rowcolor[HTML]{FFFFFF} 
  \cellcolor[HTML]{FFFFFF}                                      & \cellcolor[HTML]{FFFFFF}       & \cellcolor[HTML]{FFFFFF}                                                       & Train                         & \cellcolor[HTML]{FFFFFF}{\color[HTML]{1D1C1D} \bf{\hphantom{-}17.1 \small{{[}\hphantom{-0}8.8, 25.4{]}}}} & \cellcolor[HTML]{FFFFFF}\bf{\hphantom{-}11.7 \small{{[}\hphantom{-0}5.7, 17.7{]}}}                         \\  
  \rowcolor[HTML]{e0e0e0} 
  \cellcolor[HTML]{FFFFFF}                & \multirow{-2}{*}{\cellcolor[HTML]{FFFFFF}{\begin{tabular}[l]{@{}l@{}} \cellcolor[HTML]{FFFFFF}{$\beta_{\text{~Feature coefficients}}$} \\ \cellcolor[HTML]{FFFFFF}{\small{(from a linear student)}} \end{tabular}}} & \multirow{-2}{*}{\cellcolor[HTML]{FFFFFF}\hphantom{-}3.4 {[}-2.0, 8.8{]}} & Test                                              & {\color[HTML]{1D1C1D} \hphantom{-0}7.3 \small{{[}\hphantom{0}-0.8, 15.4{]}}*}                         & \hphantom{-0}5.0 \small{{[}\hphantom{0}-0.7, 10.7{]}}*                                                  \\ 
  \rowcolor[HTML]{FFFFFF} 
  \cellcolor[HTML]{FFFFFF}                                      & \cellcolor[HTML]{FFFFFF}       & \cellcolor[HTML]{FFFFFF}                                                       & Train                         & \cellcolor[HTML]{FFFFFF}{\color[HTML]{1D1C1D} \bf{\hphantom{-}25.6 \small{{[}\hphantom{-}17.5, 33.7{]}}}} & \cellcolor[HTML]{FFFFFF} \bf{19.1 \small{{[}\hphantom{-}13.3, 25.0{]}}}                         \\  
  \rowcolor[HTML]{e0e0e0} 
  \multirow{-8}{*}{\cellcolor[HTML]{FFFFFF}BERT}                & \multirow{-2}{*}{\cellcolor[HTML]{FFFFFF}{\begin{tabular}[l]{@{}l@{}} \cellcolor[HTML]{FFFFFF}{ $\beta_{\text{~Global cues}}$ } \\ \cellcolor[HTML]{FFFFFF}{\small{(from a linear student)}} \end{tabular}}} & \multirow{-2}{*}{\cellcolor[HTML]{FFFFFF}-1.7 {[}-6.9, 3.6{]}} & Test                                                   & {\color[HTML]{1D1C1D} \bf{\hphantom{-}18.7 \small{{[}\hphantom{-}10.8, 26.6{]}}}}                         & \bf{\hphantom{-}14.3 \small{{[}\hphantom{-0}8.7, 19.9{]}}}                                                  \\ 
 \bottomrule
  \label{tbl:something}
  \end{tabular}
  \caption{ We report the fixed effect term $\beta_{\text{treatment}}$ relative to the control for the 3 target metrics. Bootstrapped $95$\% confidence 
  intervals are in the parentheses. 
  We observe that none of the explanations help participants 
  simulate the models, %
  whereas global explanations for the BERT model and 
  feature coefficients for the logistic regression model definitively help participants  
  reduce model confidence.
  Bold values indicate 
  p-value $<0.05$ compared to the control  and $*$ 
  indicates p-value $<0.1$.
  }
  \label{tbl:mixed_effects}
  \end{table*}
  
\paragraph{Global Explanations.}
Besides local explanations, 
we experiment with global explanations that 
indicate 
common input-output associations that the models exploit.
To obtain global explanations for the BERT model,
we take inspiration from prior work 
on knowledge distillation~\cite{liu2018improving} 
to first train a linear \emph{student} model using BERT predictions 
on unseen hotel reviews.  
Since the original dataset from~\citet{ott2011finding}
contains only $1600$ reviews, 
we mine additional $13.7$K hotel reviews from TripAdvisor.\footnote{To 
download
additional reviews we 
follow a protocol similar to the data 
collection process used for the original dataset.} 
Note that we only require the BERT predictions for these reviews,
rather than the ground truth labels.
The student model achieves a simulation accuracy of 88.2\%
on the downloaded set of reviews.
We then use the trained student model to 
identify the words with the highest feature weights 
associated with both the classes.
We present the top-$20$ words for each class to participants during the training phase.
Alongside these global cue words, we also 
highlight 
words in the input as per their feature coefficients 
of the student model.
In a separate ablation study,
we isolate the effect of these global cue words 
by removing them and only highlighting input tokens 
using the feature coefficients from the student.

\subsection{Participant Details}
We recruit study participants using Amazon Mechanical Turk platform. 
We use a lightweight recruitment study that consists of $2$ 
examples (without explanations) 
to select participants.
We ask participants to guess the model prediction and edit the example to 
reduce the model confidence. 
Participants who  guess the model 
prediction within $5$ seconds 
(which we believe is insufficient to read the review) 
are filtered out.
We also remove participants who
skip the editing exercise altogether,
or 
whose edits are ungrammatical or alter the staying experience
expressed in the review.
For all our studies, we include workers 
who are residents in the United States, and have completed over $500$ HITs in the past 
and with atleast 99\% approval rate.
Workers selected from the recruitment 
test are encouraged to participate in the main study.
For the main study, 
we pay 
the workers \$20 and award a bonus of $10$ cents 
for each correct guess and
20 cents 
for every successful prediction flip. 
On an average, workers 
make 7.5 edits per review, and thus effectively see 
model predictions for 225 unique inputs. 
In total, we had $173$ participants in our main study,
with $25$ in each of the treatment and control groups (except for
one group, where $2$
participants were disqualified later for violating our instructions).
The total cost to conduct our study is about 4000 USD.

\section{Results \& Analysis}
\subsection{Do explanations help humans simulate models?}
First, %
we investigate if the
query access to 
the model's predictions and explanations
during the training phase 
enables participants to understand the models sufficiently to simulate 
its output on unseen test examples. 
We do not find evidence of improved simulatability in Tables~\ref{tbl:overall_results} and~\ref{tbl:mixed_effects}, 
where the simulation accuracy of participants---which is slightly better than random guessing---do 
not improve with access to explanations.
While prior studies~\cite{hase-bansal-2020-evaluating,chandrasekaran2018explanations}
note similar findings, 
in our opinion,
this is a stronger negative result for two reasons:
first, in our study, participants 
can alter examples and
observe model predictions and their explanations
during the training phase.  
This exercise allows participants more access to 
predictions and explanations
compared to prior studies.
Second, 
even for 
linear models, which are thought to 
be inherently ``interpretable,''
explanations do not improve simulation accuracy. 
The explanations of linear bag-of-words model
have not been examined for 
simulatability in the past.%

\subsection{Do explanations help humans perform edits that reduce the model confidence?}

Next, we examine if 
participants 
gain sufficient understanding during the training phase 
to perform edits 
that cause the models to lower the confidence 
towards the originally 
predicted class. 
Here, we find that 
logistic regression coefficient weights 
help participants 
reduce the confidence of the logistic regression model:
the average confidence reduced 
during the test phase, when they had access to explanations in training,
is 3.0 points higher than the no-explanation control. %
This difference is statistically significant with a p-value $<0.05$.
The benefits 
of such explanations during the training phase are large (over 13 points),
which is unsurprising as the faithful explanations 
 shown during the training phase can guide participants 
to effectively edit the document to lower model confidence.
During the training phase, 
they are able to statistically significantly 
flip more predictions, however, 
this ability 
does not 
transfer to the test phase.

For the BERT model,
neither LIME nor integrated gradients %
help participants flip more predictions at the test phase.
Integrated gradients-based explanations 
are effective 
only during the training phase.
In contrast, %
the feature coefficients,
from a linear student 
model 
help 
participants reduce the model confidence of the BERT model---both at train and test time, 
demonstrating
how associations
from a simple student model 
can lead 
to actionable insights about the original BERT model. 
Including global cue words 
alongside
feature coefficients 
markedly improves participant's ability to manipulate 
the BERT model. 
This fact that among all the inspected methods, 
attributions from a linear student model 
are the most effective 
emphasizes the 
need to explicitly 
evaluate explanations 
with their intended users,
instead of relying on the qualitative 
inspection of a few examples. 

Another noteworthy result here is that
we are able to quantitatively 
differentiate the effectiveness of different explanations 
using the ``percentage of examples flipped'' and ``average confidence reduced'' metrics  
from the editing exercise proposed in this paper.
This contrasts with the 
the previously used simulatability metric, 
therefore, 
we recommend future studies 
on evaluation of interpretability techniques  
to consider (similar) editing tasks and metrics instead. 
\begin{table*}
\centering
    \begin{tabular}{ll|ll|ll}
    \toprule
                    &     & \multicolumn{2}{c|}{First Edits}                                                                                                                & \multicolumn{2}{c}{All Edits}                                                                                                      \\
    \cmidrule(r){3-6}
    \multirow{-2}{*}{Model}       & \multirow{-2}{*}{Treatments}                & Deletion                                     & Substitution                                    & Deletion                                        & Substitution                                   \\
     \midrule
    \begin{tabular}[l]{@{}l@{}} Logistic \\ Regression \end{tabular}                           & Feature coefficients  & 48.5 {[}42.5, 54.6{]}      & 43.0 {[}38.3, 47.9{]}     & 40.4 {[}37.0, 43.8{]}                        & 41.8 {[}39.2, 44.6{]} \\
     \midrule
                                       & LIME                  & 56.8 {[}49.8, 63.7{]}      & 67.2 {[}62.8, 71.5{]}     & 39.2 {[}35.2, 43.2{]}                        & 49.2 {[}46.1, 52.2{]} \\
    \multirow{2}{*}{\cellcolor[HTML]{FFFFFF}BERT}                                                                           & \cellcolor[HTML]{FFFFFF}Integrated gradients  & 48.1 {[}42.0, 54.3{]}      & 50.8 {[}46.2, 55.4{]}     & 32.4 {[}29.2, 35.6{]} & 45.9 {[}43.2, 48.6{]} \\
    [-1.5ex]
   & \cline{1-5}
    \multicolumn{1}{l}{\cellcolor[HTML]{FFFFFF}}                                      & \cellcolor[HTML]{FFFFFF}{\begin{tabular}[l]{@{}l@{}} \cellcolor[HTML]{FFFFFF}{Feature coefficients} \\ \cellcolor[HTML]{FFFFFF}{\small{(from a linear student)}} \end{tabular}}  & 42.5 {[}36.6, 48.5{]}      & 47.1 {[}42.3, 52.0{]}     & 33.9 {[}30.5, 37.3{]} & 42.4 {[}40.0, 44.8{]} \\
    \bottomrule
    \end{tabular}
    \caption{The table records
    the percentage of first, and all, 
    edits performed on words that are among the 
    top 20\% highlighted words in the review. 
    Participants 
    prefer editing highlighted words, indicating that they respond to the presented explanations.
    If participants were to uniformly edit the reviews, 
    the top-$20$\% highlighted words would receive 
    about 20\% of first and all edits. 
    }
    \label{tbl:editing_preferences}
\end{table*}
\begin{figure}[h]
  \centering
    \includegraphics[width=0.75\linewidth]{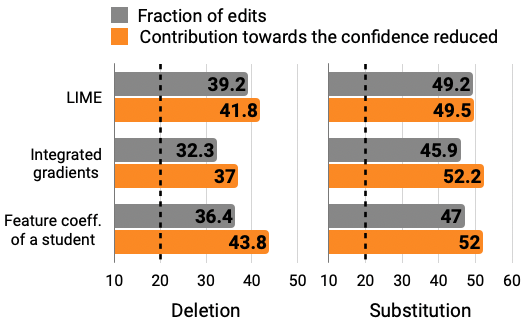}
  \caption{Percentage of edits on the top-$20$\% highlighted words and their contribution towards the confidence reduction. This plot indicates that (a) highlighted tokens receive a bulk of edits (compared to their quantity, which for the purposes of this experiment is $20$\%); and (b) edits performed on tokens highlighted via integrated gradients and feature coefficients are effective in reducing confidence.}
  \label{img:effective_edits}
\end{figure}
\subsection{Do participants edit tokens highlighted as explanations? 
Are their edits effective?}
One other benefit of our framework is that, in contrast to previous studies, it allows us to directly monitor whether participants are paying attention to the explanations, specifically by measuring how they respond to highlighted words.
To do so, we record 
the fraction  
of times edits are performed on a word that is among the top-$20$\% of highlighted words in a given input text.
If there is no preference towards highlighted words, 
this value would be close to $20$\%.
From Table~\ref{tbl:editing_preferences}, 
we see that the participants 
edit the highlighted 
words %
significantly more often, both with respect to first edits and all edits.
For instance, 
across all explanations,
for about 45-55\% of examples,
the first deleted or substituted word 
is a word among the top-$20$\% of highlighted words.

We further analyze if the edits on the top-$20$\%
highlighted words are effective in reducing model 
confidence. In Figure~\ref{img:effective_edits}, we plot the 
fraction of edits on highlighted words and their 
contribution in reducing model 
confidence. 
We compute their contribution by 
aggregating 
the fractional reduction in model confidence caused due to that edit. 
We 
inspect 
if these edits are more effective 
than those performed on the remaining words.
We find
that the edits on highlighted words are more 
effective, however, their effectiveness varies with different
explanation types. 
Edits on words highlighted using 
integrated 
gradients and feature coefficients of the student 
model 
have larger contribution 
towards reducing model confidence than edits on the words highlighted via LIME.
This result corroborates our previous findings 
suggesting that integrated gradients and feature 
coefficients from a student model are statistically significantly 
more helpful in reducing model confidence during the training phase 
\paragraph{Do people use global cues?} %
For the treatment group wherein we present 
participants 
feature coefficients and $40$ global cue words
from a linear student model as explanations for the BERT classifier, 
we determine  the 
extent to which 
participants 
use the global cues. %
We report that around
one in five edits 
utilizes %
global cues both during the training and the testing phase. 
The fraction of insertions that contain 
global cue words are 17.1\% and 18.2\% for training and testing respectively.
Further, the percentage of deletions that contain 
these cue words are 21.2\% in training and 17.3\% in the testing phase.
These results reveal
    that participants indeed incorporate global cue words while editing, 
    and as shown in~Tables~\ref{tbl:overall_results} and \ref{tbl:mixed_effects}, the edits 
    performed when these cues are present
    are effective in lowering the model confidence and flipping predictions.

\section{Conclusion}
A common argument 
for providing explanations 
is that they (ought to)
improve human's understanding about a model; 
however, 
many prior studies note that they do not improve 
their ability to simulate the model (which is primarily used as a proxy for model understanding).
In this work, we extend 
the prior evaluation paradigm by 
instead asking participants 
to edit the input
examples with an objective to  
reduce model confidence towards the predicted class. 
This exercise
allows us to compute detailed metrics, namely, the 
average confidence reduced and the percentage of examples 
flipped. 
We evaluate several explanation techniques for both a linear model 
and BERT-based classifier. 
Similar to past findings, we first note that for both these models, 
none of the considered explanations improve 
model simulatability.
We also find that participants with access to feature coefficients 
during training 
can force a larger drop in the model confidence \emph{during testing}, 
when attributions are unavailable.
Interestingly, for BERT-based classifier, 
global cue words 
and feature coefficients, obtained 
using a linear student model trained to 
mimic its predictions, 
prove to be effective. 
These results reveal that 
associations from a linear student model 
could   
provide insights for a BERT-based model, 
and importantly, the editing paradigm 
could be used to differentiate the relative utility of explanations. 
We recommend 
future studies on evaluating interpretations  
to consider similar
 metrics.

\section*{Acknowledgements}

We thank Michael Collins, Mansi Gupta, Bhuwan Dhingra and Nitish Kulkarni 
for their feedback. %
In addition, we acknowledge 
Khyathi Chandu, Aakanksha Naik, Alissa Ostapenko, Vijay Viswanathan,
Manasi Arora and Rishabh 
Joshi for painstakingly 
testing the user interface of our study.

\bibliography{aaai22.bib}

\end{document}